\def\year{2019}\relax
\documentclass[letterpaper]{article}
\usepackage{aaai19}
\usepackage{times}
\usepackage{helvet}
\usepackage{courier}
\usepackage{url} 
\usepackage{graphicx}
\graphicspath{{figures/}}
\usepackage{amsmath}
\usepackage{dsfont}
\usepackage{algorithm}
\usepackage{algorithmic}
\usepackage{booktabs}
\usepackage{multirow}
\usepackage{mathrsfs}

\title{Projection Convolutional Neural Networks\\for 1-bit CNNs via Discrete Back Propagation}
\author{Jiaxin Gu,\textsuperscript{1} Ce Li,\textsuperscript{2} Baochang Zhang,\textsuperscript{1}\thanks{Baochang Zhang is the corresponding author.} Jungong Han,\textsuperscript{3} Xianbin Cao,\textsuperscript{1} \\ 
	{\bf \Large  Jianzhuang Liu,\textsuperscript{4} David Doermann\textsuperscript{5}}  \\
\textsuperscript{1} Beihang University, 
\textsuperscript{2} China University of Mining and Technology, Beijing\\
\textsuperscript{3} Lancaster University,
\textsuperscript{4} Huawei Noah's Ark Lab,
\textsuperscript{5} University at Buffalo \\
\{jxgu1016, bczhang\}@buaa.edu.cn, celi@cumtb.edu.cn\\
}

\begin{document}
	\maketitle
	\begin{abstract}
			The advancement of deep convolutional neural networks (DCNNs) has driven significant improvement in the accuracy of recognition systems for many computer vision tasks. However, their practical applications are often restricted in resource-constrained environments. In this paper, we introduce projection convolutional neural networks (PCNNs) with a discrete back propagation via projection (DBPP) to improve the performance of binarized neural networks (BNNs). The contributions of our paper include: 1) for the first time, the  projection function is exploited to efficiently solve the discrete back propagation problem, which leads to a new highly compressed CNNs (termed PCNNs); 2) by exploiting multiple projections, we  learn a set of diverse quantized kernels that compress the full-precision kernels in a more efficient way than those proposed previously; 3) PCNNs achieve the best classification performance compared to other state-of-the-art BNNs on the ImageNet and CIFAR datasets.
	\end{abstract}
	
	\section{Introduction}
	Deep convolutional neural networks (DCNNs) have shown significant ability to learn powerful feature representations directly from image pixels. However, their success has come with requirements for large amounts of memory and computational power, and the practical applications of most DCNNs are limited on smaller embedded platforms and in mobile applications. In light of this, substantial research efforts are being invested in saving bandwidth and computational power by pruning and compressing redundant parameters generated by convolution kernels \cite{paper01,paper02}. 
	
	One typically promising method is to compress the representations via approximating floating point weights of the convolution kernels by binary values \cite{rastegari2016xnor,paper10,paper14}. Recently, Local Binary Convolution (LBC) layers have been introduced in \cite{paper15}, to approximate the non-linearly activated responses of standard convolutional layers. In \cite{paper14}, a BinaryConnect scheme using real-valued  weights as a key reference is exploited for the binarization process. In \cite{paper10,rastegari2016xnor},  XNOR-Net is presented where both the kernel weights and inputs attached to the convolution are approximated with binary values, thus allowing an efficient implementation of the convolutional operations. In \cite{zhou2016dorefa,lin2017towards}, DoReFa-Net exploits bit convolution kernels with low bitwidth parameter gradients to accelerate both training and inference. While ABC-Net \cite{lin2017towards} adopts multiple binary weights and activations to approximate full-precision weights such that the prediction accuracy degradation can be alleviated. More recently,  a simple fixed scaling method incorporated in a 1-bit convolutional layer is employed to binarize CNNs, obtaining closet-to-baseline results with minimal changes  \cite{mcdonnell2018training}. Modulated convolutional networks (MCNs) are presented in \cite{cvprxiaodi2018} to merely binarize the kernels, which achieves better results than the baselines.
		
	While reducing storage requirements greatly, these BNNs generally have significant accuracy degradation, compared to those using the full-precision kernels. This is primarily due to the following two reasons. (1) The binarization of CNNs could be essentially solved based on the discrete optimization, but it has long been neglected in previous work. Discrete optimization methods can often provide strong guarantees about the quality of the solutions they find and lead to much better performance in practice \cite{felzenszwalbdiscrete,kim2017dctm,Laude_2018_CVPR}. (2) The loss caused by the binarization of CNNs has not been well studied.  
	
	\begin{figure*}[h!]
		\centering
		\includegraphics[width=0.7\linewidth]{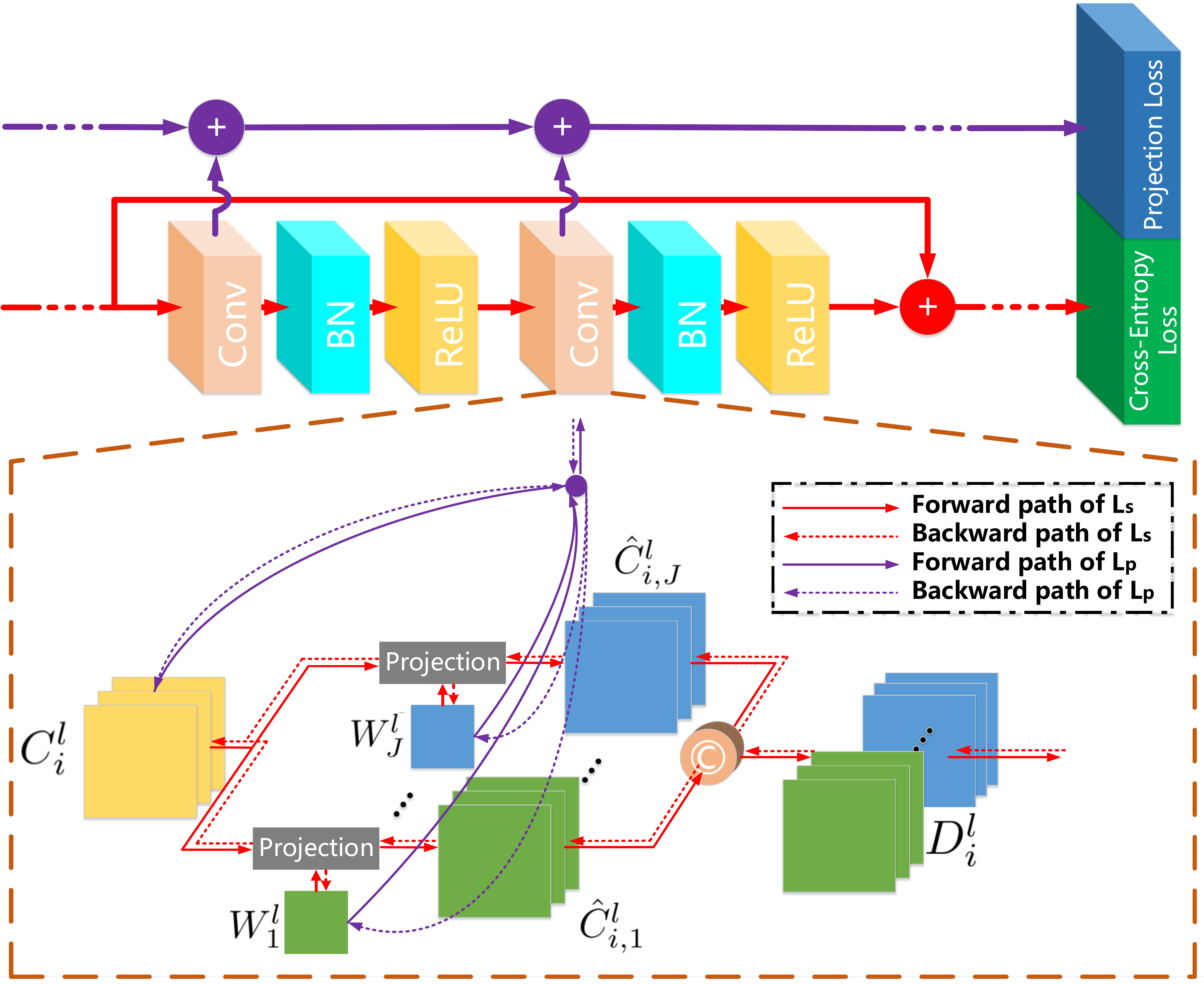}
		\caption{In PCNNs, a new discrete back propagation via projection is proposed to build  binarized neural networks in an end-to-end  manner. Full-precision convolutional kernels $C_i^l$ are quantized to be $\hat{C}_{i,j}^l$  via the projection. Due to multiple projections,  the diversity is enriched. The resulting kernel tensor $D_i^l$ is used {the} same as in conventional ones. Both the projection loss $L_p$ and the traditional loss $L_s$ are used to train PCNNs. We illustrate our network structure {\it Basic Block Unit} based on Resnet, and more specific details are shown in the dotted box (projection convolution layer). \copyright \ indicates the concatenation operation on the channels. Note that the projection matrixes $W_j^l$ {and full-precision kernels $C_i^l$} are not used in the inference.}
		\label{main-structure}
		\vspace{-13pt}
	\end{figure*}
	In this paper, we propose a new discrete back propagation via projection (DBPP) algorithm to efficiently build  our projection convolutional neural networks (PCNNs) and obtain highly-accurate yet robust BNNs. In the theoretical framework, for the first time, we achieve a projection loss by taking advantage of our DBPP algorithm, i.e., capacity of discrete optimization, on model compression. The advantages of the projection loss also  lie in that, on the one hand, it can be jointly learned with the conventional cross-entropy loss in the same pipeline as back propagation; on the other hand, it can enrich diversity and thus improve the modeling capacity of PCNNs. As shown in Fig.\ref{main-structure}, we develop a generic projection convolution layer which can be easily used in existing convolutional networks, where  both quantized kernels and the projection   are jointly optimized in an end-to-end manner. Due to the projection matrices only used for optimization but not for reference, resulting in a compact and portable learning architecture, the PCNNs model can be highly compressed and also efficient which outperforms all other state-of-the-art BNNs. The contributions of this paper include:
	
	(1) A new discrete back propagation via projection (DBPP) algorithm is proposed to build  BNNs in an end-to-end  manner. By exploiting multiple projections, we learn a set of diverse quantized kernels that thus compress the full-precision models in a better way.
	
	(2) A projection loss is theoretically achieved in DBPP, and we develop a generic projection convolutional layer to efficiently binarize existing convolutional networks, such as VGGs and Resnets.
	
	(3) PCNNs achieve the best classification performance compared to other state-of-the-art BNNs on the ImageNet and CIFAR datasets.
	
	\section{Related Work}
	Existing CNNs compression works generally follow three pathways, which are quantized neural networks (QNNs) \cite{PruningFilter,he2017channel,CVPR2018zhou,ICLR2018wu,deepcompression},  sparse connections \cite{low_rank1,low_rank2} and designing new CNN architectures \cite{mobilenet,shufflenet,squeezenet}. Recent research efforts on quantized neural networks (QNNs) have considerably reduced the memory requirement and computation complexity in DCNNs by using codebook-based network pruning \cite{PruningFilter}, Huffman encoding \cite{deepcompression}, Hash functions \cite{paper06}, low bitwidth weights and activations \cite{qnn}, and further generalized low bitwidth gradients \cite{zhou2016dorefa} and errors \cite{ICLR2018wu}. 
	BinaryNet based on BinaryConnect is proposed to train DCNNs with binary weights, where the activations are triggered at run-time while the parameters are computed at training time \cite{paper10}. The XNOR-Net is introduced to approximate the convolution operation using primarily binary operations, which reconstructs unbinarized kernels using binary kernels with a single scaling factor \cite{rastegari2016xnor}. 
	
	\begin{table*}[htbp]
		\caption{A brief description of main notation used in the paper.}
		\centering
		\begin{tabular}{l l l l}
			\toprule
			$C^l_i$: full-precision kernel & $W_j^l$: projection matrix & $D_i^l$: set of $\hat{C}^l_{i,j}$ & $\Omega$: discrete set\\
			$\hat{C}^l_{i,j}$: quantized kernel & $\widetilde W_j^l$: duplicated $W$ & $\lambda$: trade-off scaler for $L_P$ & $a_i$: discrete value in $\Omega$\\
			\hline
			$i$: kernel index & $j$: projection index & $l$: layer index & $[k]$: iteration index\\
			$I$: number of kernels & $J$: total projection number & $L$: number of layers & $h$: plane index\\
			\bottomrule
		\end{tabular}
		\label{tab:variables}
	\end{table*}
	
	Our target is similar to those binarization methods that all attempt to reduce memory consumption and replace most arithmetic operations with binary operations. However, the way we quantize the kernels is different in two respects. First, we use an efficient discrete optimization based on  projection, in which the continuous values tend to converge to a set of nearest discrete values within our framework. Second, multiple projections are introduced to bring diversity into BNNs and further improve the performance.
	
	\section{Discrete Back Propagation via Projection}
	Discrete optimization is one of hottest topics in mathematics and is widely used to solve computer vision problems \cite{kim2017dctm,Laude_2018_CVPR}. In this paper, we propose a new  discrete  back propagation algorithm, where a  projection function is exploited to binarize or quantize the input variables in a unified framework. Due to a flexible projection scheme in use, we actually obtain diverse binarized models of higher performance than previous ones. In Table \ref{tab:variables}, we describe the main notation used in the following sections.
	
	\subsection{Projection}
	In our work, we define the quantization of the input variable as a projection onto a set, 
	
	\begin{equation}
	\Omega:=\{a_1, a_2, ..., a_U\},
	\label{eq:omega}
	\end{equation}
	where each element $a_i$, $i=1,2,...,U$ satisfies the constraint $a_1<a_2<...<a_U$, and is the discrete value of the input variable. Then we define the projection of $x \in \rm R$ onto  $\Omega$ as:
	\begin{equation}
	\begin{aligned}
	P_{\Omega}(x)&= \arg \min_{a_i}\Arrowvert x-a_i\Arrowvert, i \in \{1,...,U\}, 
	\end{aligned}
	\label{projection}
	\end{equation}
	which indicates that the projection  aims to find the nearest discrete value for each continuous value $x$.
	
	\subsection{Optimization}
	{For any $f(x)$ whose gradient exists, we minimize it based on the discrete optimization method.} Conventionally, the discrete optimization problem is solved by searching for an optimal set of discrete values with respect to the minimization of a loss function. We propose that in the $k$th iteration, based on the projection in Eq. \ref{projection}, $x^{[k]}$ is quantized to $\hat{x}^{[k]}$ as:
	\begin{equation}
	\hat{x}^{[k]}=P_{\Omega}(x^{[k]}),
	\label{eq:quanti}
	\end{equation}
	which is used to define our optimization problem as:
	\begin{equation}
	\label{optimization}
	\begin{array}{cc}
	\min & f(x) \\
	\mbox{s.t.} 
	& \hat{x}_j =  P^{j}_{\Omega}(\omega_j \circ x),
	\end{array}
	\end{equation}
	where ${\omega_j}$ is a projection matrix\footnote{Since the kernel parameters $x$ are represented as a matrix, $\omega_j$ is also represented as a matrix.}, and $\circ$ denotes the Hadamard product. 
	The new minimization problem in (\ref{optimization}) is hard to solve using back propagation (e.g., deep learning paradigm)  due to the new constraint 
	$\hat{x}^{[k]}_j = P^{j}_{\Omega}(\omega_j\circ x^{[k]})$. 
	To solve the problem within the back propagation framework, we define our update rule  as:
	\begin{equation}
	x\leftarrow x^{[k]}-\eta \delta_{\hat{x}}^{[k]},
	\end{equation}
	where the superscript $[k+1]$ is dropped from $x$, $\delta_{\hat{x}}$ is the gradient of $f(x)$ with respect to $x=\hat{x}$, and $\eta$ is the learning rate. The quantization process $\hat{x}^{[k]}\leftarrow x^{[k]}$, i.e., $P^{j}_{\Omega}(\omega_j\circ x^{[k]})$, is equivilent to finding the projection of $\omega_j \circ(x+\eta\delta_{\hat{x}}^{[k]})$ onto $\Omega$ as:
	\begin{equation}
	\hat{x}^{[k]}= \arg \min_{\hat x} \lbrace \Arrowvert\hat{x}- \omega_j \circ(x+\eta\delta_{\hat{x}}^{[k]})\Arrowvert^2, \hat{x}\in \Omega \rbrace.
	\label{xhat}
	\end{equation}
	
	Obviously, $\hat{x}^{[k]}$ is the solution to the problem in (\ref{xhat}). So, by incorporating (\ref{xhat})  into $f(x)$,  we obtain a new formulation for (\ref{optimization}) based on the Lagrangian method as:
	\begin{equation}
	\min f(x)+\lambda \sum^J_j\Arrowvert\hat{x}^{[k]}- \omega_j \circ (x+\eta\delta_{\hat{x}}^{[k]})\Arrowvert^2,
	\label{lagrangian}
	\end{equation}
	where $J$ is the total number of projection matrices. The new added part (right) shown in (\ref{lagrangian}) is a quadratic function, and is referred to as the \textbf{projection loss}.
	\subsection{Projection Convolutional Neural Networks}
	Projection convolutional neural networks (PCNNs), shown in Fig. \ref{main-structure}, work by taking advantage of DBPP for model quantization. To this end, we reformulate our {projection loss}  shown in  (\ref{lagrangian}) into the deep learning paradigm as: 
	
	\begin{equation}
	L_p=\frac{\lambda}{2}\sum^{L,I}_{l,i}\sum^{J}_{j}\vert\vert \hat{C}_{i,j}^{l,[k]}-  \widetilde W_j^{l,[k]}\circ(C_i^{l,[k]}+\eta\delta_{\hat{C}_{i,j}^{l,[k]}}) \vert\vert^2,
	\label{Lp}
	\end{equation}
	where $C_i^{l,[k]}$, $l\in \{1,...,L\}, i\in \{1,...,I\}$  denotes the $i$th kernel tensor of the $l$th convolutional layer in the $k$th iteration. $\hat{C}_{i,j}^{l,[k]}$  is the quantized kernel of $C_i^{l,[k]}$ via projection $P_{\Omega}^{l,j}, j\in \{1,...,J\}$ as:
	\begin{equation}
	\hat{C}_{i,j}^{l,[k]}=P_{\Omega}^{l,j}(\widetilde W_j^{l,[k]}\circ C_i^{l,[k]}),
	\label{projection of kernels}
	\end{equation}
	where $\widetilde W_j^{l,[k]}$ is a tensor,  calculated by duplicating a learned projection matrix $W_j^{l,[k]}$ along the channels, which thus fits the dimension of $C_i^{l,[k]}$.  $\delta_{\hat{C}_{i,j}^{l,[k]}}$ is the  gradient at $\hat{C}_{i,j}^{l,[k]}$ calculated based on $L_S$, i.e., $\delta_{\hat{C}_{i,j}^{l,[k]}}=\frac{\partial L_S}{\partial \hat{C}_{i,j}^{l,[k]}}$.  The iteration index $[k]$  is omitted  in the following for simplicity.

	In PCNNs, both the cross-entropy loss and projection loss are used to build the total loss  $L$ as:
	\begin{equation}
	L = L_S+L_P.
	\label{totalloss}
	\end{equation}
	
	The proposed projection loss regularizes the continuous value converging onto $\Omega$, at the same time minimizing the cross-entropy loss, which is illustrated in Fig. \ref{figure:lambda} and Fig. \ref{figure:epoch}.
	
	\subsection{Forward Propagation based on Projection Convolution Layer}
	For each full-precision kernel $C_i^l$, the corresponding quantized kernels $\hat{C}_{i, j}^l$ are concatenated to construct the kernel $D_i^l$ which actually participates in convolution operation. 
	
	\begin{equation}
	D_i^l=\hat{C}_{i,1}^l\oplus\hat{C}_{i,2}^l\oplus\cdots\oplus\hat{C}_{i,J}^l,
	\label{Q}
	\end{equation}
	where $\oplus$ denotes the concatenation operation on tensors.
	
	In PCNNs, the projection convolution is implemented based on  $D^l$ and $F^l$  to calculate  the feature map $F^{l+1}$ of the next layer:
	\begin{equation}
	F^{l+1}=Conv2D(F^{l},D^l),
	\label{conv_1}
	\end{equation}
	where $Conv2D$ is the traditional 2D convolution. Although our convolutional kernels are 3D-shaped tensor,  we design the following strategy to fit the traditional 2D convolution:
	\begin{equation}
	F_{h,j}^{l+1}=\sum_{i,h}F_h^l \otimes D_{i,j}^l,
	\label{conv_2}
	\end{equation}
	\begin{equation}
	F_{h}^{l+1}=F_{h,1}^{l}\oplus \cdots \oplus F_{h,J}^{l}
	\label{conv_3},
	\end{equation}
	where $\otimes$ denotes the convolutional operation. $F_{h, j}^{l+1}$ is the $j$th channel of the $h$th feature map in the $(l+1)$th convolutional layer and $F_h^l$ denotes the $h$th feature map in the $l$th convolutional layer.
	
	It should be emphasized that we can utilize  multiple projections to enrich the diversity of convolutional kernels $D^l$, though the single projection already achieves much better performance. This is due to the DBPP in use, which is clearly different from  \cite{lin2017towards} in that it is based on a single quantization scheme. Within our convolutional scheme, there is no dimension disagreement on feature maps and kernels in two successive layers. Thus, we can replace the traditional convolutional layers with ours to change widely-used networks, such as VGGs and Resnets. At the inference time, we only store the set of quantized kernels $D_i^l$ instead of full-precision ones, that is, the projection matrices $W_j^l$ are not used for the inference, which achieves the reduction of storage. 
	\subsection{Backward Propagation}
	According to Eq. \ref{totalloss}, what should be learned and updated are the full-precision kernels $C_i^l$ and projection matrix $W^l$ ($\widetilde W^l$) using the updated equations described below.
	
	\subsubsection{Updating $C_i^l$:} 
	We define $\delta_{C_i}$ as the gradient of the full-precision kernel $C_i$, and have:
	\begin{equation}
	\delta_{C_i^l}=\frac{\partial L}{\partial C_i^l}=\frac{\partial L_S}{\partial C_i^l}+\frac{\partial L_P}{\partial C_i^l},
	\label{delta_c}
	\end{equation}
	\begin{equation}
	C_i^l\leftarrow C_i^l-\eta_1 \delta_{C_i^l},
	\label{updateC}
	\end{equation}
	where $\eta_1$ is the learning rate for the convolutional kernels.
		
	More specifically, for each item in Eq. \ref{delta_c}, we have:
	\begin{equation}
	\begin{aligned}
	\frac{\partial L_S}{\partial C_i^l}&=\sum^J_j \frac{\partial L_S}{\partial \hat{C}_{i,j}^l} \frac{\partial P_{\Omega}^{l,j}(\widetilde W_j^{l}\circ C_i^l)}{ \partial (\widetilde W_j^{l}\circ C_i^l)}\frac{ \partial(\widetilde W_j^{l}\circ C_i^l)}{\partial C_i^l}\\
	&=\sum^J_j \frac{\partial L_S}{\partial \hat{C}_{i,j}^l}\circ\mathds{1}	_{-1\leq\widetilde W_j^{l}\circ C_i^l\leq1}\circ\widetilde W_j^{l},
	\end{aligned}
	\label{Ls/C}
	\end{equation}
	\begin{equation}
	\frac{\partial L_P}{\partial C_i^l}=\lambda \sum^J_j \left[\widetilde W_j^{l}\circ\left(C_i^l+\eta\delta_{\hat{C}_{i,j}^l}\right)-\hat{C}_{i,j}^l\right]\circ\widetilde W_j^{l},
	\label{Lm/C}
	\end{equation}
	where $\mathds{1}$ is the indicator function \cite{rastegari2016xnor} widely used to estimate the gradient of non-differentiable function.
	\subsubsection{Updating $W_j^l$:} Likewise, the gradient of the projection parameter $\delta_{W_j^l}$ consists of the following two parts:
	\begin{equation}
	\delta_{W_j^l}=\frac{\partial L}{\partial W_j^l}=\frac{\partial L_S}{\partial W_j^l}+\frac{\partial L_P}{\partial W_j^l},
	\label{delta_w}
	\end{equation}
	\begin{equation}
	W_j^l\leftarrow W_j^l-\eta_2 \delta_{W_j^l},
	\label{updateW}
	\end{equation}
	where $\eta_2$ is the learning rate for $W_j^l$. We further have:
	\begin{equation}
		\begin{aligned}
		\frac{\partial L_S}{\partial W_j^l}&=\sum^J_h\left(\frac{\partial L_S}{\partial {\widetilde W_j^{l}}}\right)_h\\
		&=\sum^J_h\left(\sum^I_i\frac{\partial L_S}{\partial \hat{C}_{i,j}^l} \frac{\partial P_{\Omega}^{l,j}(\widetilde W_j^{l}\circ C_i^l)}{\partial(\widetilde W_j^{l}\circ C_i^l)}\frac{\partial(\widetilde W_j^{l}\circ C_i^l)}{\partial \widetilde W_j^l}\right)_h\\
		&=\sum^J_h \left(\sum^I_i\frac{\partial L_S}{\partial \hat{C}_{i,j}^l}\circ\mathds{1}_{-1\leq\widetilde W_j^{l}\circ C_i^l\leq1}\circ C_i^l\right)_h,
		\end{aligned}
		\label{Ls/w}
	\end{equation}
	\begin{equation}
		\small\frac{\partial L_P}{\partial W_j^l}\!=\!\lambda\!\sum^J_h\! \left(\! \sum^I_i \!\left[\!\widetilde W_j^{l}\circ \!\left(\!C_i^l\!+\!\eta\delta_{\hat{C}_{i,j}^l}\!\right)\!-\!\hat{C}_{i,j}^l\!\right]\!\circ \!\left(\!C_i^l\!+\!\eta\delta_{\hat{C}_{i,j}^l}\!\right)\!\right)_h,
		\label{Lm/w}
	\end{equation}
	where $h$ indicates the $h$th plane of the tensor along the channels. It shows that the proposed algorithm can be trainable in an end-to-end manner, and we summarize the training procedure in Alg. \ref{alg:algorithm}. In  implementation, we use the mean of $W$ in the forward process, but keep the original $W$ in the backward propagation.
	
	Note that in  PCNNs for BNNs, we set $U$=2 and $a_2$=$-a_1$. Two binarization processes are used in PCNNs. The first one is  the kernel binarization, which is done based on the projection onto $\Omega$, whose elements are calculated based on  the mean of absolute values of all the full-precision kernels per layer \cite{rastegari2016xnor} as:
	\begin{equation}
	\frac{1}{I}\sum^I_i \left(\|C_i^l\|_{1}\right).
	\label{a_i}
	\end{equation}
	where $I$ is the total number of kernels.	
	
	\begin{algorithm}[tb] 
		\small
		\caption{Discrete Back Propagation via Projection} 
		\label{alg:algorithm} 
		\begin{algorithmic}[1] 
			\REQUIRE ~~\\ 
			The training dataset; the full-precision kernels $C$; the projection matrix $W$; the learning rates $\eta_1$ and $\eta_2$.
			\ENSURE ~~\\ 
			The PCNNs based on the updated $C$ and $W$.
			\STATE Initialize $C$ and $W$ randomly; 
			\REPEAT
			\STATE // Forward propagation
			\FOR {$l=1$ to $L$} 
			\STATE $\hat{C}_{i,j}^l\leftarrow Project(C_i^l)$; // using Eq. \ref{projection of kernels}
			\STATE $D_i^l\leftarrow Concatenate(\hat{C}_{i,j})$; // using Eq. \ref{Q}
			\STATE Perform activation binarization; //using the sign function
			\STATE Traditional 2D convolution; // using Eq. \ref{conv_1}, \ref{conv_2} and \ref{conv_3}
			\ENDFOR
			\STATE // Backward propagation
			\STATE Compute $\delta_{\hat{C}_{i,j}^l}=\frac{\partial L_S}{\partial \hat{C}_{i,j}^l}$;
			\FOR {$l=L$ to $1$} 
			\STATE // Calculate the gradients
			\STATE calculate $\delta_{C_i^l}$; // using Eq. \ref{delta_c}, \ref{Ls/C} and \ref{Lm/C}
			\STATE calculate $\delta_{W_j^l}$; // using Eq. \ref{delta_w}, \ref{Ls/w} and \ref{Lm/w}
			\STATE // Update the parameters
			\STATE $C_i^l\leftarrow C_i^l-\eta_1 \delta_{C_i^l}$; // Eq. \ref{updateC} 
			\STATE $W_j^l\leftarrow W_j^l-\eta_2 \delta_{W_j^l}$; //Eq. \ref{updateW}
			\ENDFOR
			\STATE Adjust the learning rates $\eta_1$ and $\eta_2$.
			\UNTIL {the network converges}
		\end{algorithmic}
	\end{algorithm}
	
	\section{Experiments}
	PCNNs are evaluated  on the object classification task with CIFAR10/100 \cite{krizhevsky2014cifar} and ILSVRC12 ImageNet  datasets \cite{deng2009imagenet}. Our DBPP algorithm can be applied to any DCNNs. For fair comparison with other state-of-the-art BNNs, we use Wide-Resnet (WRN) \cite{zagoruyko2016wide}, Resnet18 \cite{he2016deep} and VGG16 \cite{simonyan2014very} as our full-precision backbone networks, to build our PCNNs by replacing their full-precision convolution layers  with our projection convolution.
	
	\subsection{{Datasets} and Implementation details}
	\subsubsection{{Datasets:}}
	CIFAR 10/100 \cite{krizhevsky2014cifar} are  natural image classification datasets containing a training set of 50K and a testing set of 10K $32 \times 32$ color images across the 10/100 classes. For CIFAR10/100 and parameter study, we employ WRNs to evaluate our PCNNs and report the accuracies. Unlike CIFAR 10/100, ILSVRC12 ImageNet object classification dataset \cite{deng2009imagenet} is more challenging due to its large scale and greater diversity. There are 1000 classes and 1.2 million training images and 50k validation images in it. For comparison of our method to the state-of-the-art on the ImageNet dataset, we adopt Resnet18 and VGG16 to validate the superiority and effectiveness of PCNNs.
	\begin{figure}[tb]
		\centering
		\includegraphics[width=1.03\linewidth]{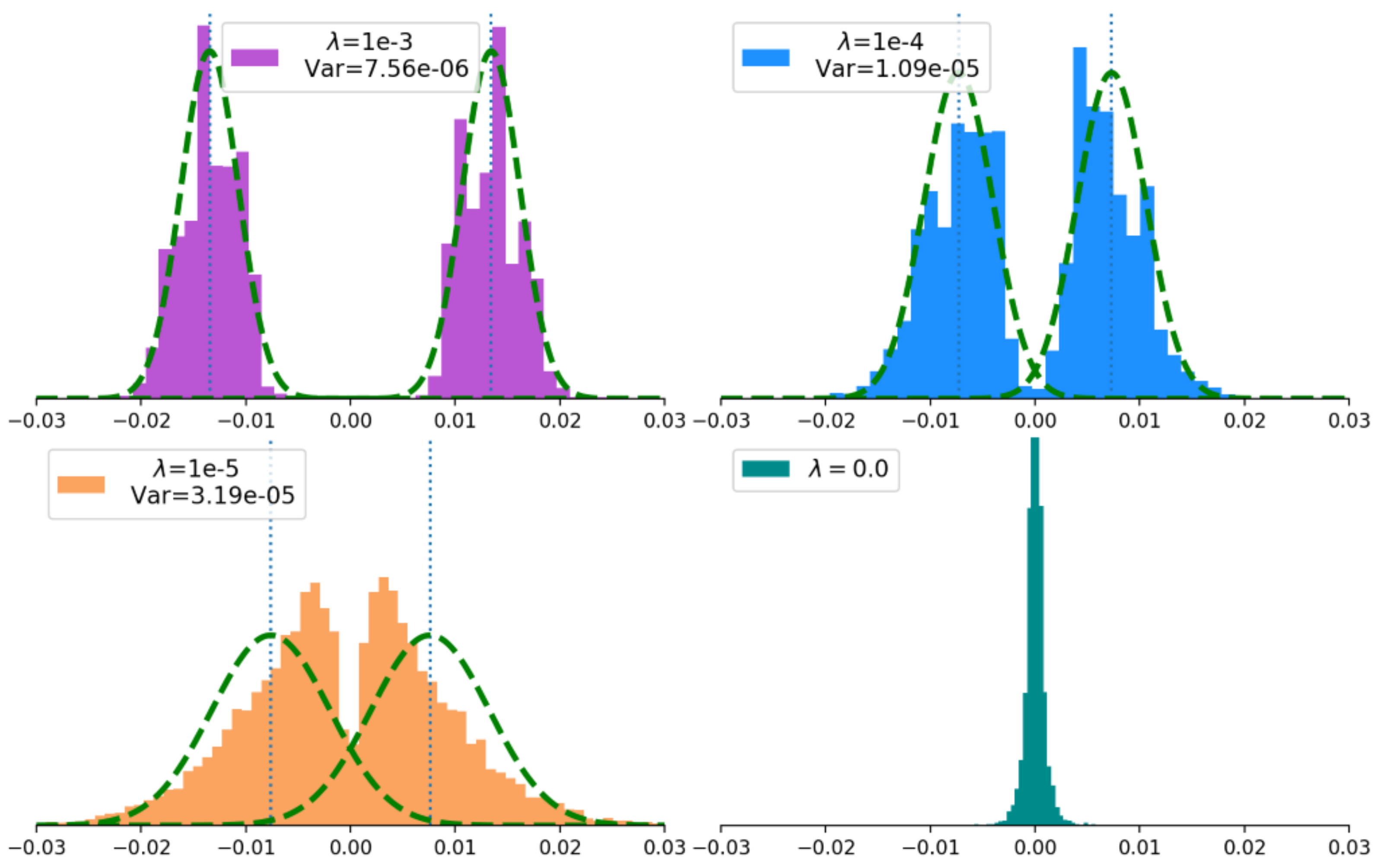}
		\caption{We visualize the kernel weights distribution of the first convolution layer of PCNN-22. When decreasing $\lambda$, which balances the projection loss and cross-entropy loss, the variance becomes larger. Particularly, when $\lambda$=0 (no projection loss),  only one cluster is obtained, wherein the kernel weights distribute  around $0$, which could result in instability during binarization. Conversely, two Gaussians (with the projection loss, $\lambda>0$) are more powerful  than the single one (without the projection loss), which thus results in better BNNs as also validated in Table \ref{table:lambda}. }  
		\label{figure:lambda}
	\end{figure}
	\begin{figure}[tb]
		\centering
		\includegraphics[width=1.03\linewidth]{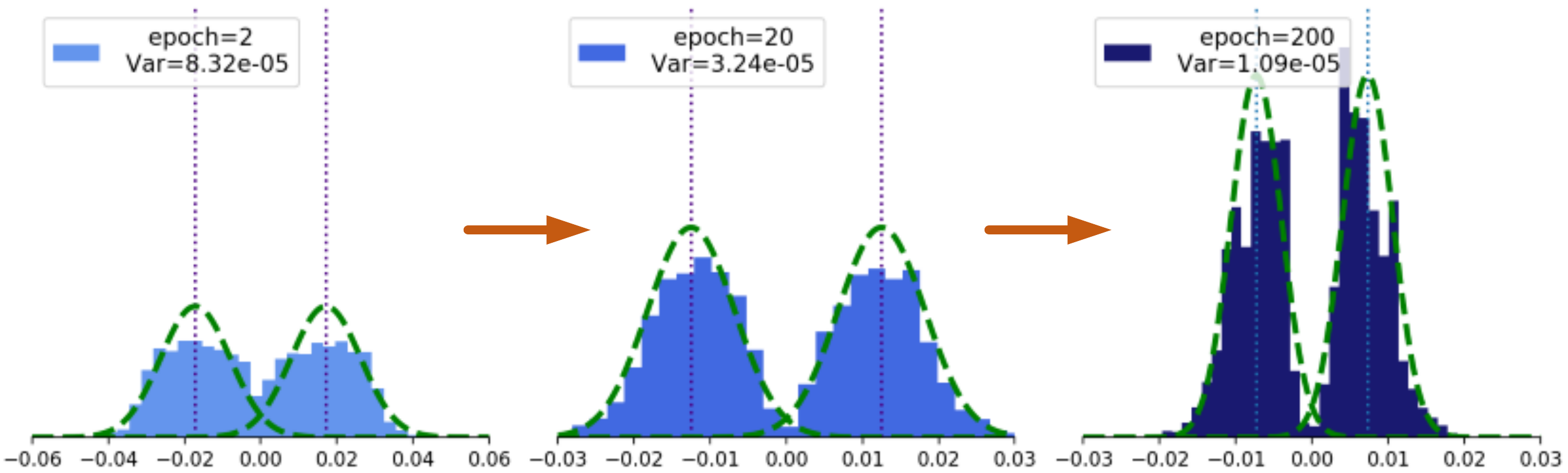}
		\caption{ With $\lambda$ fixed to $1e-4$, the variance of the kernel weights becomes smaller from the $2$th epoch to the $200$th epoch, which confirms that the projection loss does not affect the convergence.}
		\label{figure:epoch}
	\end{figure}
	
	\subsubsection{WRN:} WRN is a network structure similar to Resnet with a depth factor $k$ to control the feature map depth dimension expansion through 3 stages, within which the dimensions remain unchanged. For simplicity we fix the depth factor to $1$. Each WRN has a parameter $i$ which indicates the channel dimension of the first stage and we vary it between 16 and 64 leading to two network structures, 16-16-32-64 and 64-64-128-256. Other setting including training details is the same as \cite{zagoruyko2016wide} except that we only add dropout layers with a ratio 0.3 to the 64-64-128-256 structure in case of overfitting. 
	WRN-22 indicates a network with 22 convolutional layers and similarly for WRN-40.
	
	\subsubsection{Resnet18 and VGG16:} For these two networks, we simply replace the convolutional layers with the projection convolution layer and keep other components unchanged. To train the two networks, we choose SGD as the optimization algorithm with a momentum of 0.9 and a weight decay $1e-4$. The initial learning rate for $W_j^l$ is 0.01 and for $C_i^l$ and other parameters the initial learning rates are 0.1, with a degradation of 10\% for every 20 epochs before it reaches the maximum epoch of 60.

	\subsection{Ablation study}
	\subsubsection{Parameter:} As mentioned above, the proposed projection loss has the ability to control the process of quantization, similar to clustering. We compute the distributions of the full-precision kernels and visualize the results in Figs. \ref{figure:lambda} and \ref{figure:epoch}. The hyper-parameter $\lambda$ is designed to balance the projection loss and the cross-entropy loss. We vary it from $1e-3$ to $1e-5$ and finally set it to $0$ in Fig. \ref{figure:lambda}, where the variance becomes larger as decreasing $\lambda$. When $\lambda$=0,  only one cluster is obtained, where the kernel weights distribute tightly around the threshold=0. This could result in instability during binarization, because little noise may cause a positive weight to be negative and vice versa.
	\begin{figure}[tb]
		\centering
		\includegraphics[width=0.8\linewidth]{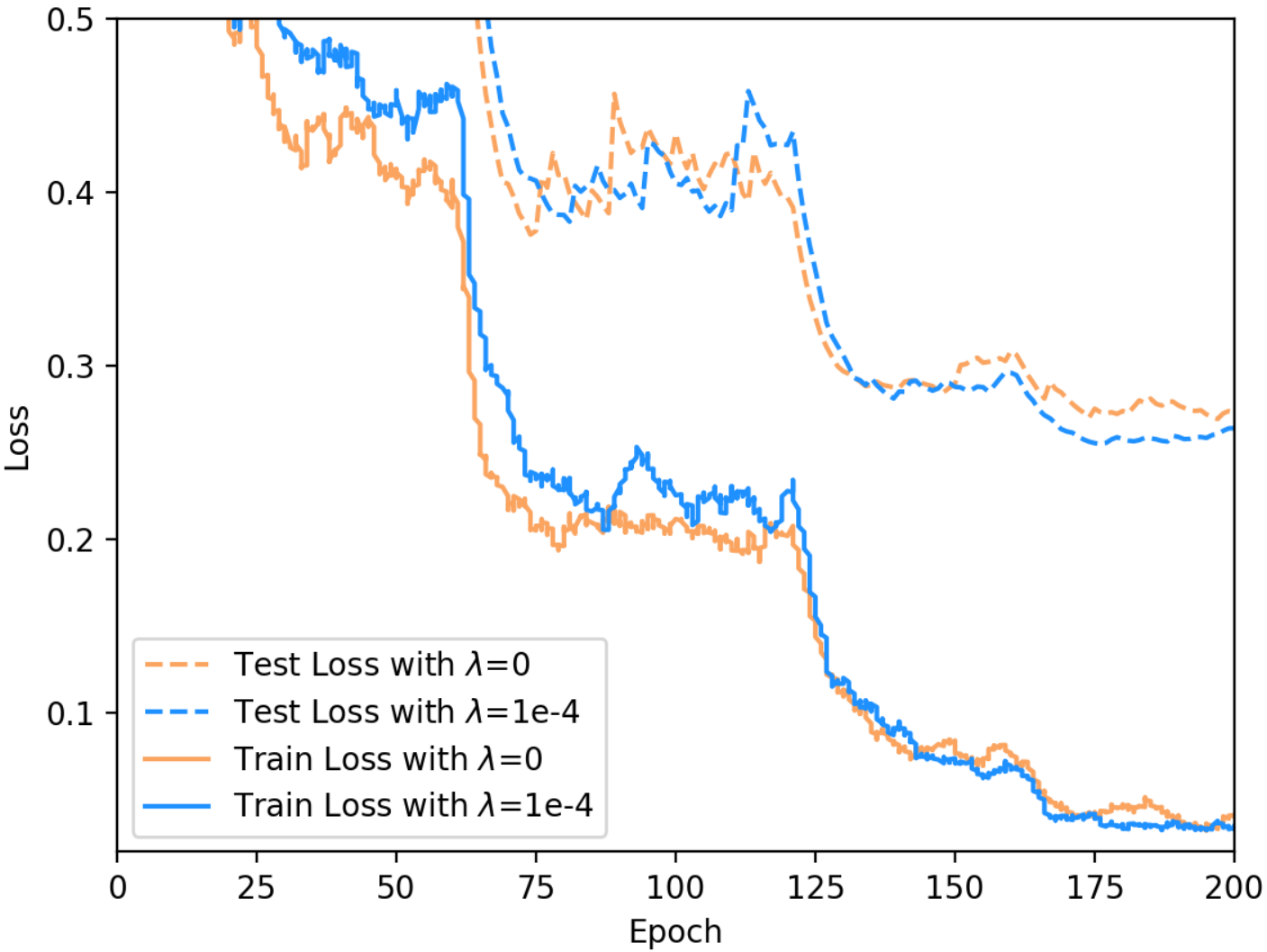}
		\caption{Training and testing curves of PCNN-22 when $\lambda$=0 and $1e-4$, which shows that the projection affects little on the convergence.}
		\label{figure:losscurve}
	\end{figure}
	
	We also show the evolution of the distribution about how the projection loss works in the training process in Fig. \ref{figure:epoch}. A natural question is: do we always need a large $\lambda$? As a discrete optimization problem, the answer is no and the experiment in Table \ref{table:lambda} can verify it, i.e., both the projection loss and cross-entropy loss should be considered at the same time with a good balance. For example, when $\lambda$ is set to $1e-4$, the accuracy is higher than {those with }other values. Thus, we fix  $\lambda$ to $1e-4$ in the following experiments.
	
	\subsubsection{Learning convergence:} For PCNN-22 in Table \ref{table:lambda}, the PCNNs model is trained for 200 epochs and then used to conduct inference. In Fig. \ref{figure:losscurve}, we plot the training and testing loss with $\lambda$=0 and $\lambda$=1e-4, respectively. It clearly shows that PCNNs with $\lambda$=1e-4 (blue curves) converge faster than PCNNs with $\lambda$=0 (yellow curves) when the epoch number $>150$.
	
	\subsubsection{Diversity visualization:} 
\begin{figure}[tb]
		\centering
		\includegraphics[width=\linewidth]{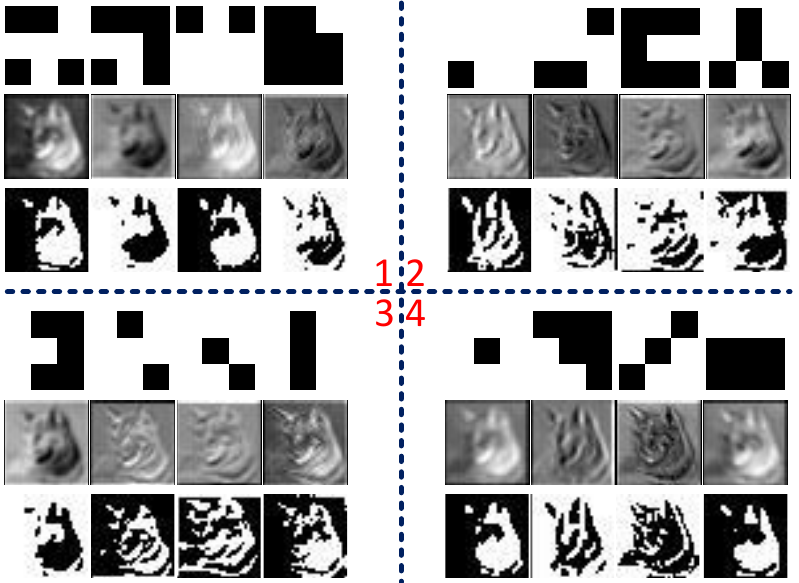}
		\caption{Illustration of binary kernels $D_i^l$ (first row), feature maps produced by $D_i^l$ (second row), and corresponding feature maps after binarization (third row) when $J$=4. This confirms the diversity in PCNNs.}
		\label{figure:diversity}
	\end{figure}

In Fig. \ref{figure:diversity}, we visualize four channels of the binary kernels $D_i^l$ in the first row, feature maps produced by $D_i^l$ in the second row, and corresponding feature maps after binarization in the third row when $J$=4, which illustrates the diversity of kernels and feature maps in PCNNs. Thus, the multiple projection functions can capture the diverse information and result in a high performance based on the compressed models
	
	\begin{table}[t]
		\caption{With different $\lambda$, the accuracy of PCNN-22 and PCNN-40 based on WRN-22 and WRN-40, respectively, on CIFAR10 dataset.}
		\centering
		\begin{tabular}{c c c c c}
			\toprule
			\multirow[c]{2}{*}{Model} & \multicolumn{4}{c}{$\lambda$} \\
			\cmidrule(lr){2-5}
			& $1e-3$ & $1e-4$ & $1e-5$ & $0$ \\
			\hline
			\hline
			PCNN-22 & 91.92 & 92.79 & 92.24 & 91.52 \\
			PCNN-40 & 92.85 & 93.78 & 93.65 & 92.84 \\
			\bottomrule
		\end{tabular}
		\label{table:lambda}
	\end{table}
	
	\begin{table}[tb]
		\caption{Test accuracy on CIFAR10/100 dataset. PCNNs are based on WRN-22. The numbers of parameters refer to the models on CIFAR10.}
		\centering
		\small
		\begin{tabular}{c c c c}
			\toprule
			\multirow[c]{2}{*}{\large Model} & \multirow[c]{2}{*}{\large \#Para.} & \multicolumn{2}{c}{Dataset} \\
			\cmidrule(lr){3-4}
			&& CIFAR-10 & CIFAR-100 \\
			\hline
			\hline
			WRN ($i$=16) & 0.27M & 92.62 & 68.83 \\
			WRN ($i$=64) & 4.29M & 95.75 & 77.34 \\
			\hline
			BinaryConnect & 14.02M & 91.73 & - \\
			BNN & 14.02M & 89.85 & -  \\
			LBCNN & 14.02M & 92.99 & - \\
			BWN & 14.02M & 90.12 & -\\
			XNOR-Net & 14.02M & 89.83 & - \\
			{\cite{mcdonnell2018training}} & 4.30M
 & 93.87 &76.13\\
			\hline
			PCNN ($i$=16, $J$=1)& 0.27M & 89.17  & 62.66 \\
			PCNN ($i$=16, $J$=2)& 0.54M & 91.27 & 66.86 \\
			PCNN ($i$=16, $J$=4)& 1.07M & 92.79 & 70.09 \\
			PCNN ($i$=64, $J$=1)& 4.29M & 94.31 & 76.93 \\
			PCNN ($i$=64, $J$=4)& 17.16M & 95.39 & 78.13 \\
			\bottomrule
		\end{tabular}
		\label{table:cifar}
	\end{table}
	
	\subsection{Results on CIFAR-10/100 datasets}
	We first compare our PCNNs with the original WRNs with initial channel dimension $i$=16 and 64 in Table \ref{table:cifar}. We compare the performance of different models with similar parameter amount. Thanks to the multiple projections in use ($J$=4), our results on both  datasets are comparable with the full-precision networks. Then, we compare our results with other state-of-the-arts such as BinaryConnect \cite{paper14}, BNN \cite{paper10}, LBCNN \cite{paper15}, BWN, XNOR-Net \cite{rastegari2016xnor} {and the model in \cite{mcdonnell2018training}}. It is observed that at least a 1.5\% accuracy improvement is gained with our PCNNs, and in most cases large margins are achieved, which indicates that  DBPP is really effective for the task of model compression. As shown in Table \ref{table:cifar}, the increase of $J$  can also boost the performance and exhibits a better modeling ability than others by the flexible projection scheme with more feature diversity gain. However, for small models like PCNN ($i$=16), the increase of $J$ seems more significant for avoiding accuracy degradation, but for large models like PCNN ($i$=64), it results in relatively small improvement. So we suggest to use small $J$=1 to avoid additional computation in large models, like the PCNNs based on Resnet18 and VGG16, which can still outperform the others.
	
	\subsection{Results on ILSVRC12 ImageNet classification dataset}
	
	\begin{table}[tbp]
		\caption{Test accuracy on ImageNet. 'W' and 'A' refer to the weight and activation bitwidth respectively. The first two PCNNs are based on Resnet18, while the last one is based on VGG16. $\dag$ and $\ddag$ indicate $J$=1 and $J$=2 respectively.}
		\centering
		\small
		\begin{tabular}{c c c c c}
			\toprule
			Model & W & A & Top-1 & Top-5 \\
			\hline
			\hline
			Resnet18 & 32 & 32 & 69.3 & 89.2 \\
			BWN & 1 & 32 & 60.8 & 83.0 \\
			DoReFa-Net & 1 & 4 & 59.2 & 81.5 \\
			XNOR-Net & 1 & 1 & 51.2 & 73.2 \\
			ABC-Net & 1 & 1 & 42.7 & 67.6 \\
			BNN & 1 & 1 & 42.2 & 67.1 \\
			Bi-Real Net & 1 & 1 & 56.4 & 79.5 \\
			\hline
			PCNN & 1 & 32 & 63.5$^{\dag}$, 66.1$^{\ddag}$ & 85.1$^{\dag}$, 86.7$^{\ddag}$ \\
			PCNN & 1 & 1 & \textbf{57.3}$^{\dag}$ & \textbf{80.0}$^{\dag}$ \\
			\hline
			\hline
			VGG16	 & 32 & 32 & 73.0 & 91.2 \\
			\hline
			PCNN & 1 & 32 & 69.0$^{\dag}$ & 89.1$^{\dag}$ \\	
			\bottomrule
		\end{tabular}
		\label{table:imagenet}
	\end{table}
	
	For the ImageNet dataset, we employ two data augmentation techniques sequentially: 1) randomly cropping patches of $224\times224$ from the original image, and 2) horizontally flipping the extracted patches in the training. While in the testing, the Top-1 and Top-5 accuracies on the validation set with single center crop are measured. We modify the architecture of Resnet18 following \cite{liu2018bi} with additional PReLU \cite{he2015delving} and the final results of our PCNNs are finetuned based on the pretrained models with only kernel weights binarized, halving the learning rate in the training.
	
	In Table \ref{table:imagenet}, we compare our PCNNs with several other state-of-the-art models. The first part of the comparison is based on Resnet18 with  69.3\% Top-1 accuracy on the full-precision model. Although  BWN \cite{rastegari2016xnor} and DoReFa-Net \cite{zhou2016dorefa} achieve  Top-1 accuracy with degradation of less than 10\%, it should be noted that they apply full-precision and 4-bit activation respectively. With both of the weights and activations binarized, the BNN model in \cite{paper10}, ABC-Net \cite{lin2017towards} and XNOR-Net \cite{rastegari2016xnor} fail to maintain the accuracy and are inferior to our PCNN. For example, compared with the result of  XNOR-Net, PCNN increases the Top-1 accuracy by 6.1\%. For a fair comparison, we set the activation of our PCNNs to full-precision and vary $J$ from 1 to 2, and  these two results consistently outperform BWN. We also compare our method with the full-precision VGG16 model, and the accuracy drop is tolerable if only weights are binarized (see the last two rows). Note that our DBPP algorithm still works very well on the large dataset, particularly when $J$=1, which further validates the significance of our method. 
	
	In short, we achieved a new state-of-the-art performance compared to other BNNs, and  much closer performance to full-precision models in the extensive experiments, which clearly validate the superiority of DBPP  for the BNNs calculation.
	
	\subsection{Memory Usage and Efficiency Analysis}
	\begin{table}[t]
		\caption{Memory usage and efficiency of convolution comparison with XNOR-Net, full-precision Resnet18, and PCNN ($J$=1). PCNN is based on Resnet18.}
		\centering
		\small
		\begin{tabular}{c c c c}
			\toprule
			Model & Memory usage & Memory saving & Speedup\\
			\hline
			\hline
			PCNN & 33.7 Mbit & 11.10 $\times$ & 58 $\times$\\
			XNOR-Net & 33.7 Mbit & 11.10 $\times$ & 58 $\times$\\
			Resnet18 & 374.1 Mbit & - & -\\
			\bottomrule
		\end{tabular}
		\label{table:memory}
	\end{table}
	
	Memory use is analyzed by comparing our approach with the state-of-the-art XNOR-Net \cite{rastegari2016xnor} and the corresponding full-precision network. The memory usage is computed as the summation of $32$ bits times the number of full-precision kernels and 1 bit times the number of the binary kernels in the networks. As shown in Table \ref{table:memory}, our proposed PCNNs, along with XNOR-Net, reduces the memory usage by $11.10$ times compared with the full-precision Resnet18. Note that when $J$ is set to $1$, the parameter amount of our model is almost the same as XNOR-Net. The reason is that the projection parameters $W_j^l$ are only used when training for enriching the diversity in PCNNs, whereas they are not used when inference. For efficiency analysis, if all of the operands of the convolutions are binary, then the convolutions can be estimated by XNOR and bitcounting operations \cite{paper10}, which gains $58\times$ speedup in CPUs \cite{rastegari2016xnor}. For $J>1$, the memory usage and computation cost for convolution are linear to $J$.
	
	\section{Conclusion and future work}
	We have proposed an efficient discrete back propagation via projection (DBPP) algorithm to obtain our projection convolutional neural networks (PCNNs), which can significantly reduce the storage requirement for computationally limited devices. PCNNs have shown to obtain much better  performance than other state-of-the-art BNNs on ImageNet and CIFAR datasets. As a general convolutional layer, the PCNNs model can also be used in other deep models and different tasks, which will be explored in our future work.
	
	\bibliography{bibliography}
\end{document}